\pdfoutput=1

\documentclass[11pt]{article}

\usepackage[]{EMNLP2023}

\usepackage{times}
\usepackage{latexsym}
\usepackage{tabularx}
\usepackage{tcolorbox}
\usepackage{colortbl}
\usepackage{tikz,tcolorbox}

\tcbuselibrary{most}

\newtcolorbox{mybox}[2][]
{colback = black!5!white, 
 colframe = black!75!black, 
 fonttitle = \bfseries,
 colbacktitle = black!85!black, 
 enhanced,
 attach boxed title to top left={yshift=-2mm,xshift=4mm},
 title=#2,#1}

\usepackage[T1]{fontenc}

\usepackage[utf8]{inputenc}

\usepackage{microtype}

\usepackage{inconsolata}
\usepackage{inconsolata}
\usepackage{makecell}
\usepackage{graphicx}
\usepackage{float}
\usepackage{subfigure}
\usepackage{multirow}
\usepackage{amsmath}
\usepackage{booktabs}
\usepackage{bm}
\usepackage{color}
\usepackage{arydshln}

\usepackage{multirow}
\usepackage{graphicx}

\title{An Automatic Evaluation Framework for Multi-turn Medical Consultations Capabilities of Large Language Models}
\author{Yusheng Liao$^{1\dagger}$, Yutong Meng$^{1\dagger}$\footnotemark[\value{footnote}], Hongcheng Liu$^{1}$, Yanfeng Wang$^{12}$, Yu Wang$^{12*}$ \\
  $^{1}$Cooperative Medianet Innovation Center, Shanghai Jiao Tong University \\
  $^{2}$Shanghai Artificial Intelligence Laboratory \\
}
\begin{document}
\maketitle
\renewcommand{\thefootnote}{\fnsymbol{footnote}}
\footnotetext[2]{Equal contribution}
\footnotetext[1]{Corresponding author}
\renewcommand{\thefootnote}{\arabic{footnote}}
\begin{abstract}

Large language models (LLMs) have achieved significant success in interacting with human. However, recent studies have revealed that these models often suffer from hallucinations, leading to overly confident but incorrect judgments. This limits their application in the medical domain, where tasks require the utmost accuracy.
This paper introduces an automated evaluation framework that assesses the practical capabilities of LLMs as virtual doctors during multi-turn consultations. Consultation tasks are designed to require LLMs to be aware of what they do not know, to inquire about missing medical information from patients, and to ultimately make diagnoses. To evaluate the performance of LLMs for these tasks, a benchmark is proposed by reformulating medical multiple-choice questions from the United States Medical Licensing Examinations (USMLE), and comprehensive evaluation metrics are developed and evaluated on three constructed test sets. A medical consultation training set is further constructed to improve the consultation ability of LLMs. The results of the experiments show that fine-tuning with the training set can alleviate hallucinations and improve LLMs' performance on the proposed benchmark. Extensive experiments and ablation studies are conducted to validate the effectiveness and robustness of the proposed framework.

\end{abstract}
\section{Introduction}

Recently, large language models (LLMs) have achieved remarkable success in following natural language instructions and performing real-world tasks \citep{elmohamed, DBLP:journals/corr/abs-2303-08774}. LLMs have demonstrated their ability to tackle complex problems and have shown immense potential to revolutionize various industries. However, despite their impressive capabilities, LLMs often suffer from a significant drawback known as hallucination. This means they tend to provide erroneous judgments and incorrect information with high confidence, as noted in previous studies \citep{DBLP:journals/corr/abs-2305-13534, DBLP:journals/corr/abs-2304-10513}. This limitation hinders their application in tasks that require utmost accuracy, such as the medical field.

\begin{figure}[t]
\centering
\includegraphics[width=1.0\linewidth]{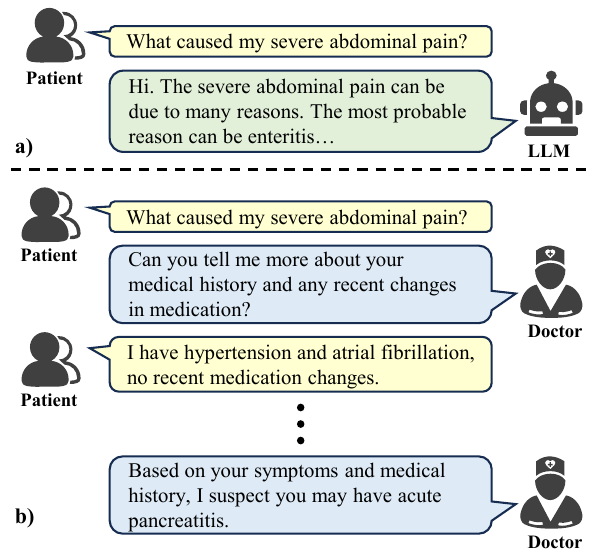}
\caption{Difference between a) the LLM and b) Doctor. The LLM, with its limitations, tends to make judgments without consulting all available information, while the doctor conducts consultations before making a final diagnosis.}
\label{fig:barplot}
\end{figure}

In clinical diagnosis, doctors typically begin with limited patient information and engage in iterative rounds of consultation to gather pertinent medical details until they reach a final diagnosis~\citep{DBLP:journals/corr/abs-2305-15075,DBLP:conf/acl/WeiLPTCHWD18}. However, LLMs with hallucinations tend to make judgments directly and depend on limited information, leading to incorrect decisions. This disparity in behavior presents a significant challenge when incorporating LLMs into the medical field, where precise diagnosis and decision-making are paramount. To truly harness the potential of LLMs for medical use, it is essential to measure their hallucinations and explore their practical abilities in multi-turn consultation conversations. Although there are already many works attempting to apply LLMs in the medical domain, most of them primarily focus on the question-answering task to explore the medical knowledge of LLMs~\citep{10.1371/journal.pdig.0000198, DBLP:journals/corr/abs-2305-09617,DBLP:journals/corr/abs-2303-13375,hiesinger2023almanac}, rather than exploring their practical abilities as a doctor.


In this paper, we introduce an automated evaluation framework for assessing the practical capabilities of LLMs as virtual doctors during multi-turn consultations. This task requires LLMs to recognize the absence of medical information, pose relevant questions throughout the consultation process, and ultimately accomplish specific objectives based on the consultation history, such as providing accurate diagnoses or offering appropriate medical advice. Therefore, LLMs that exhibit superior performance in multi-turn consultations are expected to have reduced hallucinations and enhanced medical capabilities. To implement the proposed pipeline, we begin by constructing a benchmark that decomposes the multiple-choice question into the patient's medical information and the final task to be completed by the LLMs. Then we adopt ChatGPT~\citep{elmohamed} to simulate the patient's dialogue with the doctor model during the consultation process and provide medical information that helps LLMs make a diagnosis. Lastly, we devise metrics to evaluate the performance of LLMs in multi-turn consultation dialogues by considering indicators for patients, doctors, and final tasks. In summary, the major contributions of our paper are summarized as follows:

\begin{itemize}
\setlength{\itemsep}{0pt}
    \item We propose a benchmark for multi-turn consultation in LLMs, which assesses their practical capabilities as virtual doctors. This is the first such benchmark to our knowledge and can measure the hallucination and medical ability of LLMs.
    \item We develop comprehensive evaluation metrics for scoring multi-turn conversations. Extensive experiments validate the reliability of the LLM-based metrics and their correlation with traditional metrics.
     \item We improve the consultation ability and alleviate the hallucination of LLMs by constructing a training set through reformulating multiple choice questions.
\end{itemize}

\section{Related Works}


There are already plenty of works attempting to adopt LLMs in the medical domain to act as a doctor~\cite{DBLP:journals/corr/abs-2305-15075, DBLP:journals/corr/abs-2303-14070, DBLP:journals/corr/abs-2304-08247}. Compared to the traditional medical language model like BioGPT~\cite{DBLP:journals/bib/LuoSXQZPL22}, medical LLMs fine-tuned on instruction have a stronger ability to interact with humans and show the potential to deploy in the real world. However, existing works found that it is difficult for LLMs to ensure the accuracy of generated content, which limits its further application. Therefore, it is essential to propose an evaluation method to measure the medical ability of the LLMs. Lots of medical benchmarks only focus on measuring the medical knowledge~\cite{10.1371/journal.pdig.0000198, DBLP:journals/corr/abs-2305-09617,DBLP:journals/corr/abs-2303-13375,hiesinger2023almanac} or the response ability~\cite{DBLP:conf/emnlp/ZengYJYWZZZDZFZ20,DBLP:conf/nlpcc/LiuTCLZL22} of the models, instead of exploring their practical abilities as a doctor. 

Our proposed automatic evaluation pipeline has the following innovative points compared to previous work: 1) Different from previous works only generate one response according to the dialog history, our pipeline can conduct real multiple rounds of conversation test. The proposed evaluation pipeline uses ChatGPT to simulate patients which can generate responses to the LLMs at each turn during the consultation process. 2) LLMs are required to complete the entire clinical diagnosis process from consultation to decision-making, which can measure the medical capabilities of models comprehensively.

\section{Proposed Framework}
In this section, we will first introduce the data structure and task formulation. Then we will provide a detailed description of the automatic pipeline for evaluating the LLMs. Finally, we will discuss the construction of the dataset.

\subsection{Task Formulation}
Each instance of the data consists of two parts, a patient's medical information $M$ and a task $T$ which is required to be completed at the end of the conversations. Since the consultations typically begin with the question of the patients, we also generate a fixed patient's initial request $P_0$ for each instance based on its medical information. $P_0$ generally includes a rough description of the symptoms and the requests for help. As a result, the dataset with size equal to $N$ can be formulated as:
\begin{equation}
    \{(M^{(i)}, P^{(i)}_0, T^{(i)})\}_{i=1}^{N}
\end{equation}

The whole evaluation pipeline consists of two LLMs, the doctor LLM and the patient LLM. These two are responsible for being evaluated and interacting with the doctor LLM under certain information, respectively. At the beginning of the conversation, doctor LLM can get a preliminary understanding of the patient's situation and need according to the $P_0$. Then it is required to ask questions to get useful information from the patient in the following communication. Specifically, in the $i$-th round of conversation, the doctor LLM give an query $D_i=\{d^{(i)}_1, d^{(i)}_2,...,d^{(i)}_{|D_i|}\}$ based on the conversation history in an autoregressive manner:
\begin{equation}\small
    p_{\theta}(D_i|D_{1:i-1}, P_{0:i-1}) = \prod^{|D_i|}_{k=1}p_{\theta}(d^{(i)}_k|D_{1:i-1}, P_{0:i-1})
\end{equation}
where $|D_i|$ is the length of the doctor LLM $i$-th outputs and $\theta$ is the parameter of the doctor LLM. Similarly, patient LLM  generates a response $P_i=\{p^{(i)}_1, p^{(i)}_2,...,p^{(i)}_{|P_i|}\}$ based on the query of the doctor, conversation history, and medical information:
\begin{equation}\small
    p_{\phi}(P_i|D_{1:i}, P_{0:i-1}, M) = \prod^{|P_i|}_{k=1}p_{\theta}(p^{(i)}_k|D_{1:i}, P_{0:i-1}, M)
\end{equation}
where $\phi$ represents the parameters of the patient LLM. When the doctor is no longer inquiring about patient information, we consider the consultation process to be over. At the end of the consultation conversations, doctor LLMs are requested to complete some tasks according to the whole conversation, such as providing diagnosis, etiology, or treatment plan, etc. 

\begin{figure}[tbp]
	\centering  
	\begin{mybox}[colback=white, fontupper=\small]{Prompt for Patient LLM}
    Marv is a patient that interacts with doctor to get medical help. Marv's condition can be summarized as \texttt{\{\{medical\_info\}\}}. \\
    At each Round Marv responds in less than 15 words. \\
    At each Round Marv gives only one point of information. \\
    Marv always talks colloquially. \\
    \\
    \texttt{\{\{dialog\_history\}\}} \\ 
    Marv:
    \end{mybox}
\caption{The prompt for patient LLMs. Elements in double braces $\{\{\}\}$ are replaced with question-specific values.}
\label{fig:patient prompt}
\end{figure}

\subsection{Patient Simulation}

Traditional evaluation of consultations is difficult when it comes to assessing a model's ability to handle multi-round conversations, which are inherently uncertain and uncontrollable. However, the emergence of large dialogue models has made automatic multi-round conversations possible. These models have undergone instruction fine-tuning, resulting in strong interactive abilities that can simulate the process of multiple rounds of dialogue. Additionally, large models can generate output according to specified requirements. In this paper, we utilize large-language models with instruction fine-tuning to simulate the behavior of patients in consultation conversations, thus allowing for an easy evaluation of the doctor model's ability in multiple rounds of consultations.

\renewcommand{\dblfloatpagefraction}{.9}
\begin{figure*}[thbp]
\centering
\includegraphics[width=1.0\linewidth]{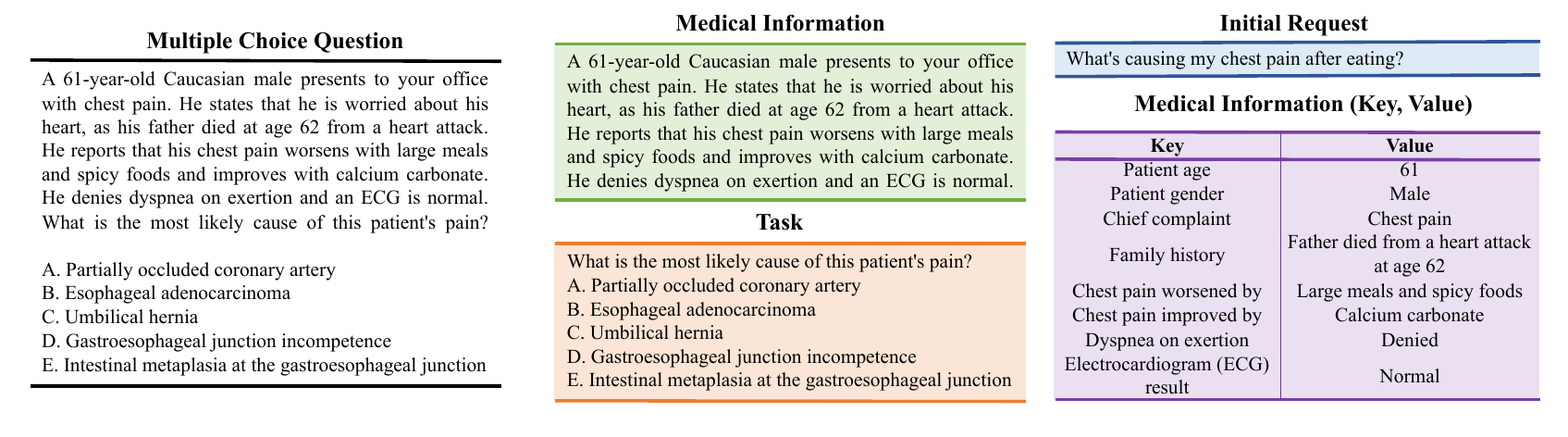}
\caption{Example of the constructed data. The multiple choice question is first split into medical information and task part. Then the initial request and key-value pair list are extracted by ChatGPT from the medical information.}
\label{fig: data_case}
\end{figure*}

\begin{figure*}[thbp]
 \centering  
 \subfigbottomskip=2pt 
 \subfigcapskip=-5pt 
 
 \subfigure[MedQA]{
  \includegraphics[width=0.3\linewidth]{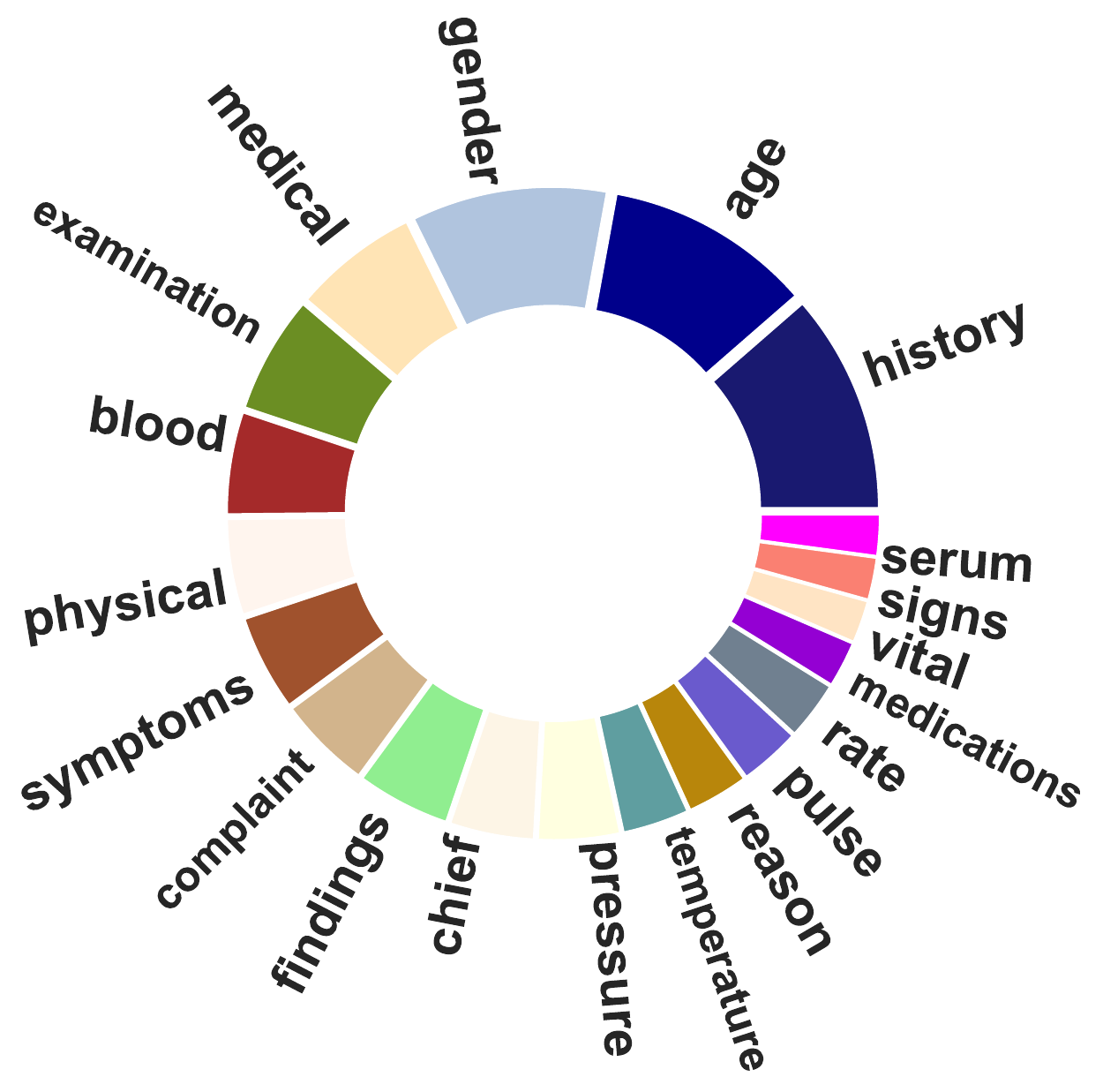}}
  \subfigure[QMax]{
  \includegraphics[width=0.3\linewidth]{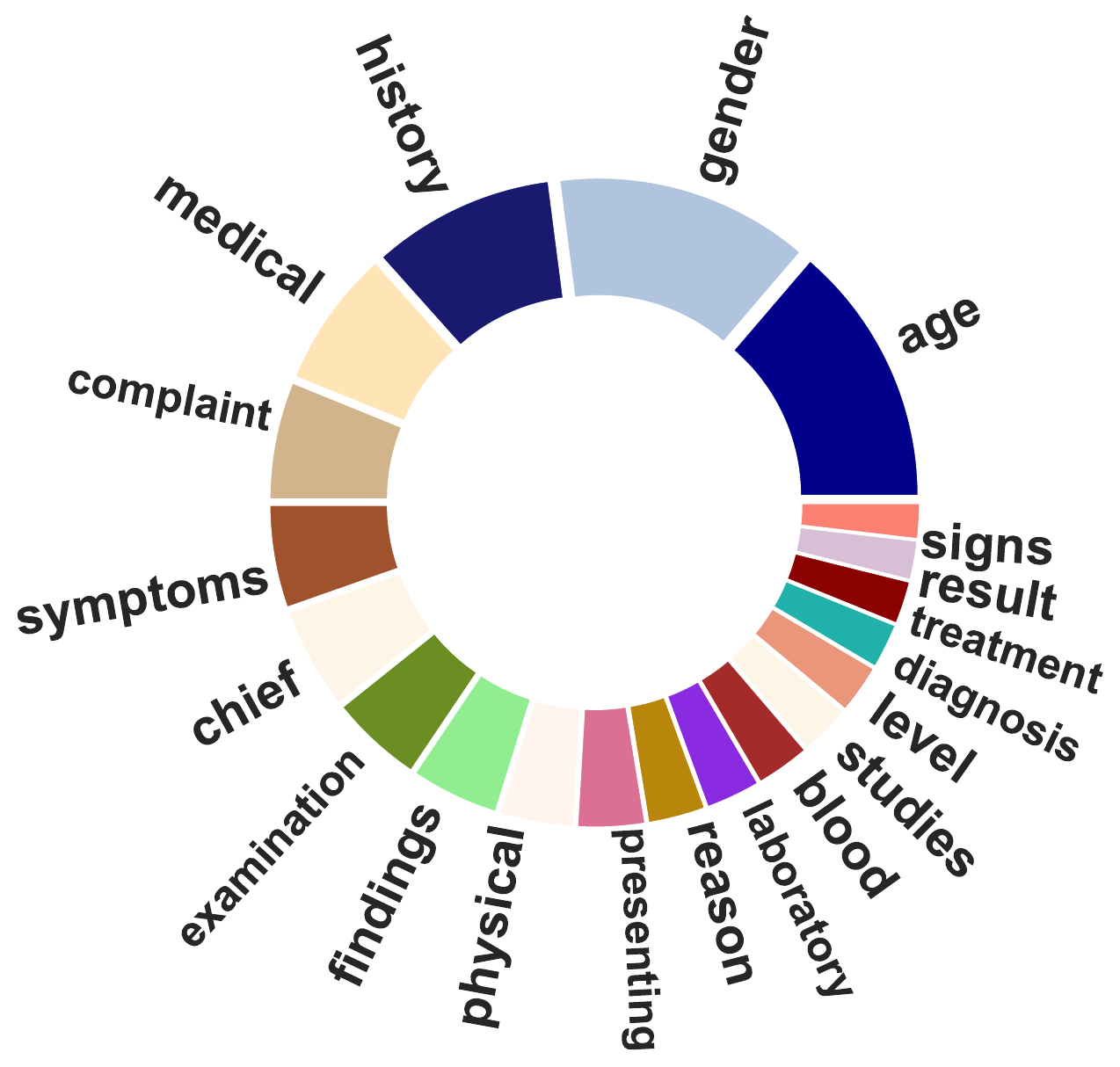}}
  \subfigure[Sample Exam]{
  \includegraphics[width=0.3\linewidth]{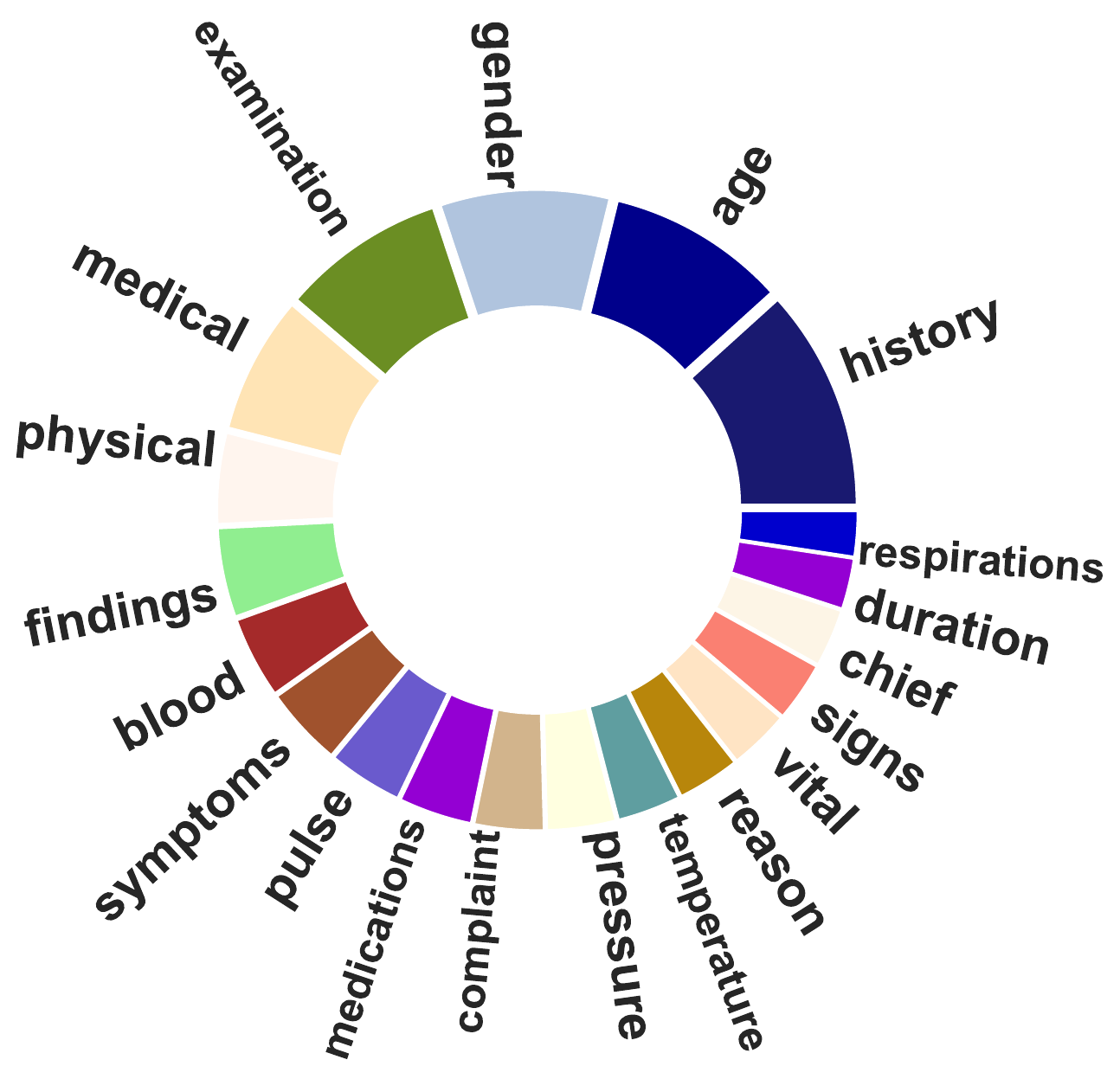}}

\caption{Most frequent words distribution of three datasets. We draw the Top-20 most frequent keys of extracted medical information. Note that the same word is represented by the same color across different datasets.}
\label{fig: high-frequency words}
\end{figure*}

We have defined some fundamental requirements for the patient LLM to ensure the robustness of the pipeline:
\begin{itemize}
\setlength{\itemsep}{0pt}
    \item \textbf{Honesty}. The patient LLM should provide an accurate and rational description of the symptoms in medical information, without reporting any nonexistent symptoms. This allows the doctor LLM to obtain correct information and reduces uncertainty in multiple rounds of dialogue.
    \item \textbf{Strictness}. As a comprehensive patient simulator, the patient model is provided with all information in ground truth. However, experiments have shown that this can lead to the model revealing medical information that doctors have not inquired about, resulting in the leakage of extra information. To be in line with real-world patients, the patient model is supposed to give information in a "lazy" mode, thereby improving the discriminability of the pipeline towards the doctor LLMs.
    \item \textbf{Colloquialism}. To better align with reality, we assume that patients do not have much medical knowledge, so their output should be colloquial. In experiments, we have also found that the colloquial prompt construction helps with the strictness requirements, which is also reported in \citet{DBLP:journals/corr/abs-2305-13614}.
\end{itemize}

In summary, we designed a prompt for the patient that meets the three characteristics described above (as shown in Figure~\ref{fig:patient prompt}). This allowed us to obtain a large language model that can better simulate the patient.

\begin{table}[tbp]
  \centering
  
  \resizebox{1.0\linewidth}{!}{
    \begin{tabular}{lccc}
    \toprule
    \multirow{2}[0]{*}{\textbf{Statistics}} & \multirow{2}[0]{*}{\textbf{MedQA}} & \multirow{2}[0]{*}{\textbf{QMax}} & \textbf{Sample} \\
    ~ & ~ & ~ & \textbf{Exam}\\
    \midrule
    \# of instances & 1127 & 1483 & 299 \\
    \# of options per question & 5.00 & 5.11 & 5.12 \\
    \# of tokens per medical information & 106.00 & 80.85 & 119.28 \\
    \# of tokens per initial request & 14.01 & 14.28 & 14.32 \\
    \# of items per medical information & 7.87 & 7.81 & 8.25 \\
    \bottomrule
    \end{tabular}
    }
    \caption{The statistics of MedQA, QMAX, Sample Exam Datasets}
\label{tab:statistic result}
\end{table}%

\subsection{Data Construction}
\label{sec:data construction}
Because there is a scarcity of data that provides comprehensive medical information and diagnostic results, we reformulated the multiple choice questions in physician exams as the test datasets for our benchmark. We have created three datasets and filtered out questions that do not contain patient information:
\begin{itemize}
\setlength{\itemsep}{0pt}
    \item \textbf{MedQA}: We have chosen the English test set of the MedQA\footnote{https://github.com/jind11/MedQA}, which was collected from Medical Licensing Examinations of the United States~(USMLE).
    \item \textbf{QMax}: The dataset includes materials purchased from the USMLE-RX\footnote{https://usmle-rx.com} resources. There are a total of 11 categories, such as cardiology, endocrine, gastroenteric, etc.
    \item \textbf{Sample Exam}: Sample exam materials were sourced from USMLE practice materials\footnote{https://www.usmle.org/prepare-your-exam}. We collect the dataset from the provided by~\citet{10.1371/journal.pdig.0000198}. In addition, we have removed the duplicated question in the test set.
\end{itemize}


For each dataset, we first split the multi-choice question into two parts: the informational part $M$ and the task part $T$. These respectively describe the patient background and task requirements. Then, we use ChatGPT to generate the patient's initial request at the beginning of the consultation conversation, based on the medical information. We also extract the main medical information into key-value pairs, following the approach proposed by \citet{DBLP:conf/emnlp/AgrawalHLKS22}.
Figure~\ref{fig: data_case} provides an example of the multi-choice question and its corresponding preprocessed data. Additionally, we calculate the most frequently occurring words in the keys of the medical information for the three datasets, as shown in Figure~\ref{fig: high-frequency words}. Detailed statistical information for the three datasets is also presented in Table~\ref{tab:statistic result}.
\section{Experiments}



\subsection{Automated Metrics}
Considering that LLMs have achieved great success in language fluency, our evaluation mainly focuses on the ability of doctor LLMs in multi-turn dialogue consultation. We introduce several indicators to measure the performance of the doctor LLMs in different dimensions.

\paragraph{Medical Information Coverage Rate} We use three types of ROUGE scores~\cite{lin-2004-rouge} to measure the amount of medical information presented in dialogue at the word level: Recall, Precision, and F1 scores. We evaluate the entire conversation at a fine-grained level from both the doctor's and patient's perspectives. For the patient evaluation, we compute three types of ROUGE-1 scores between the output of the patient LLM and the value of the extracted medical information $M$. The Recall represents how much information is exposed by the patient LLM, the Precision reflects the information density output by the patient model, and the F1 score provides a comprehensive evaluation of the patient LLM output. For the doctor evaluation, we measure the similarity between the output of the doctor LLM and the key of the extracted medical information. Specifically, the Recall measures the hallucination level of doctor LLMs. The Precision represents the professionalism of doctor LLM, as medical information generally contains medical terms. The higher the proportion of medical terms in the output of the doctor's model, the more professional the model's output we suppose. The F1 score is also a comprehensive evaluation of doctor LLMs.


\paragraph{Accuracy of the Final Task} After the multi-turn consultation, the model is required to complete a specific task, which here is to choose the right answer for a multiple-choice question according to the consultation dialogue.  The accuracy of this task reflects the accuracy and coverage of medical information throughout the entire consultation process, which is an important indicator of the consultation ability of the doctor LLM. Considering that most LLMs perform poorly on the question of the USMLE 

We use ChatGPT-Turbo-3.5 as our final task solver due to its accessibility and robustness.

\paragraph{Average Turn and Average Length} To evaluate the performance of different models in multi-turn consultations, we need to measure the average number of rounds and the average length of each conversation. The doctor LLM will stop asking questions once it assumes that the medical information is enough to make a decision. Average Turn is an indicator measuring both doctor LLMs precision and hallucination level as more hallucinated models tend to stop asking questions at earlier rounds. Ideally, we hope that models with higher accuracy in consultations also require fewer rounds and shorter conversations.


\begin{table*}[thbp]
\resizebox{\linewidth}{!}{%
\begin{tabular}{cccccccccccc}
\toprule
\multirow{2}{*}{\textbf{Dataset}} & \multirow{2}{*}{\textbf{Case}} & \multicolumn{4}{c}{\textbf{Patient}} & \multicolumn{4}{c}{\textbf{Doctor}} & \multirow{2}{*}{\textbf{Turn}} & \multirow{2}{*}{\textbf{Acc.}} \\
 &  & \textbf{Rec.} & \textbf{Pre.} & \textbf{F1} & \textbf{Len.} & \textbf{Rec.} & \textbf{Pre.} & \textbf{F1} & \textbf{Len.} &  &  \\
 \midrule
\multirow{6}{*}{MedQA} & Upper-B & - & - & - & - & - & - & - & - & - & 53.50 \\
 & Lower-B & 3.45 & 22.78 & 5.75 & - & - & - & - & - & - & 35.49 \\
 & ChatGPT-L & 28.26 & 17.84 & 20.65 & 9.15 & 18.67 & 3.58 & 5.82 & 15.94 & 9.17 & 39.31 \\
 & ChatGPT-S & 19.74 & 21.45 & 19.17 & 10.65 & 18.06 & 4.08 & 6.27 & 30.29 & 4.58 & 37.27 \\
 & Vicuna & 22.91 & 18.58 & 18.42 & 8.60 & 11.11 & 3.90 & 5.34 & 9.82 & 8.22 & 38.60 \\
 & Vicuna-FT & \textbf{33.19} & \textbf{23.55} & \textbf{26.02} & 9.17 & \textbf{23.06} & \textbf{6.11} & \textbf{9.02} & 14.77 & 8.32 & \textbf{40.20} \\
 \midrule
\multirow{6}{*}{QMAX} & Upper-B & - & - & - & - & - & - & - & - & - & 55.02 \\
 & Lower-B & 5.13 & 21.87 & 8.00 & - & - & - & - & - & - & 36.95 \\
 & ChatGPT-L & 32.55 & 12.94 & 17.69 & 9.19 & 19.82 & 2.77 & 4.72 & 16.16 & 9.24 & 41.94 \\
 & ChatGPT-S & 27.96 & \textbf{16.84} & 19.82 & 10.97 & \textbf{21.98} & 3.17 & 5.29 & 37.95 & 4.14 & 41.40 \\
 & Vicuna & 26.72 & 14.44 & 16.55 & 7.63 & 11.28 & 2.97 & 4.37 & 9.88 & 8.04 & 39.45 \\
 & Vicuna-FT & \textbf{33.17} & 16.62 & \textbf{20.79} & 8.86 & 21.62 & \textbf{4.83} & \textbf{7.30} & 14.26 & 7.84 & \textbf{42.75} \\
 \midrule
\multirow{6}{*}{Sample Exam} & Upper-B & - & - & - & - & - & - & - & - & - & 60.20 \\
 & Lower-B & 3.51 & 23.69 & 5.83 & - & - & - & - & - & - & 40.13 \\
 & ChatGPT-L & 26.98 & 18.70 & 20.71 & 9.32 & 19.35 & 4.17 & 6.66 & 16.54 & 9.12 & 43.81 \\
 & ChatGPT-S & 19.74 & 23.28 & 19.52 & 10.97 & 18.70 & 4.65 & 6.93 & 31.56 & 4.43 & 45.15 \\
 & Vicuna & 23.30 & 21.01 & 19.95 & 8.55 & 11.87 & 4.60 & 6.12 & 10.28 & 8.27 & 42.81 \\
 & Vicuna-FT & \textbf{32.59} & \textbf{24.43} & \textbf{26.05} & 9.42 & \textbf{22.21} & \textbf{6.85} & \textbf{9.70} & 13.82 & 8.27 & \textbf{46.82}\\
\bottomrule
\end{tabular}%
}

\caption{Performance of different models on three datasets. Len and Turn represent the average length of the sequence in each turn and the average number of turns for each consultation instance. The best results in each dataset are bold.}
\label{tab: main result}
\end{table*}

\subsection{Doctor Construction} \label{section: doctor construction}
To test the robustness of our framework, we employed several LLMs to simulate doctors and test various prompts for each LLM.


\paragraph{Upper Bound} In this case, the model can directly complete the final task according to the golden medical information. The accuracy of the upper bound is limited by the medical ability of the model for task completion.

\paragraph{Lower Bound} In the lower bound, the models were required to complete the final task only based on the initial request of the patient. This case can be used to indicate how much medical information is contained in the initial request. Therefore, the incremental information during the consultation process can be measured by the increase in accuracy.

\paragraph{ChatGPT-Turbo-3.5}
In general, ChatGPT does not actively ask questions. Therefore, we prompted ChatGPT to act like a doctor in order to complete the proposed tasks in the evaluation pipeline. In the experiments, we observed that ChatGPT would like to draw a conclusion after a brief inquiry, where the number of rounds of consultation is generally less than five. Therefore, we designed two versions of Doctors with different prompts named ChatGPT-Long and ChatGPT-Short, respectively. The long version tries to guide the model to ask as many questions as possible, up to 10 questions, and the short version doesn't require the model to ask excessive questions. In both settings, the model will naturally come to a conclusion when it thinks the information provided is enough to answer the patient's initial question.

\paragraph{Vicuna}
Vicuna is an open-sourced LLM trained by fine-tuning LLaMA~\citep{DBLP:journals/corr/abs-2302-13971} on user-shared conversations collected from ShareGPT\footnote{https://sharegpt.com/}, and has demonstrated strong multi-round interaction ability. We used the same prompt as ChatGPT-Short for Vicuna to act like a doctor. 

\paragraph{Vicuna Finetuned}
We created training data using the training set of the MedQA dataset to illustrate Vicuna's consultation abilities. First, we reconstructed the main medical information in key-value pairs, which is the same process used for the test data. Next, we generated a golden consultation dialogue between a doctor and a patient, with ChatGPT playing the role of both. In each turn, the doctor ChatGPT asked a key question based on the medical information, and the patient ChatGPT answered it. Finally, the doctor ChatGPT provided an analysis and diagnosis.

\begin{figure*}[!h]
\centering
\includegraphics[width=1\linewidth]{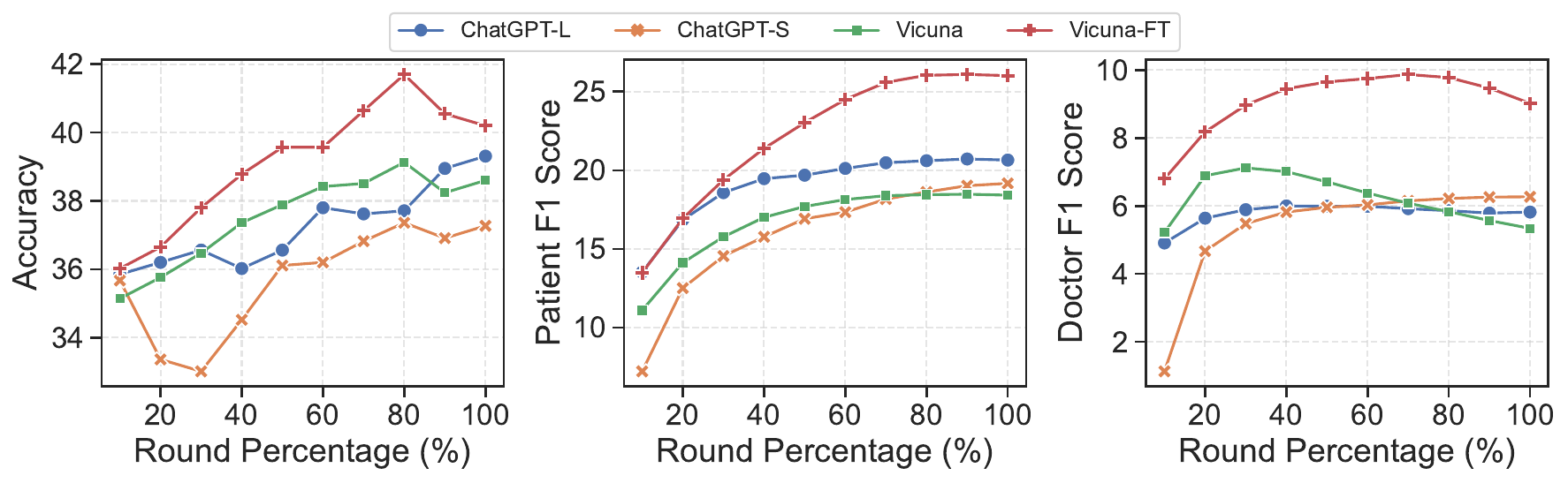}

\caption{Turn Analysis results of different LLMs on MedQA dataset.}
\label{fig:turn_analysis}
\end{figure*}

\subsection{Main Result}
\label{sec: main result} 
Table~\ref{tab: main result} presents the results of various models discussed in Section~\ref{section: doctor construction} on three datasets. The accuracy gap between the Upper-Bound and Lower-Bound cases is significant, highlighting how insufficient medical information often leads to degraded diagnostic performance of the models. After several rounds of interaction with patient models, the accuracies improve in all cases, indicating that current models possess basic consultation abilities and can deduce missing information to some extent.

Comparing two versions of ChatGPT, we observe that ChatGPT-Long requires approximately twice as many consultation rounds as ChatGPT-Short, while the length of its questions is about half that of ChatGPT-Short. Despite the different prompts eliciting different behaviors, both models exhibit similar performance according to the patients/doctor F1 scores and task accuracy. This suggests that the model's ability is the primary factor determining its effectiveness, rather than the design of the prompts.

Although Vicuna performs relatively poorly in the consultation process compared to ChatGPT, it demonstrates much stronger capabilities after being fine-tuned on the constructed training data. Vicuna-FT achieves the best F1-score on all datasets while keeping the average lengths of questions and answers short. This shows that Vicuna-FT has less hallucination problem and are more precise and professional. With such powerful inquiry ability, the model can obtain more detailed medical information from patients, thereby improving the accuracy of the final task. In fact, it can even surpass ChatGPT by up to 1.67 percent.




\section{Analysis}
In this section, we conduct more experiments to analyze the robustness of the proposed framework. We first decompose the consultation process and explore the performance of different models at the turn level. Then we assess the diversity of the models' outputs and the impact of the consultation order on the final accuracy. Finally, we investigate the models' performance on cases with different complexity.

\subsection{Turn Number Analysis}
\label{section: percentage}
We conducted an investigation into how medical information was revealed throughout the multi-turn dialogue. Because different doctor LLMs have very different distributions of turn numbers, we used the percentage of rounds finished as the index for dialogue completeness. Specifically, for the case where the percentage of rounds is equal to 20$\%$, if the total number of rounds is 10, the first two rounds of the conversation will be selected as the consultation process. We plot important metrics, including accuracy and F1-scores of patient LLM and doctor LLM in Figure~\ref{fig:turn_analysis}.

\begin{table}[tbp]
\resizebox{\columnwidth}{!}{%
\begin{tabular}{cccc}
\toprule
\textbf{Model} & \textbf{ROUGE-1} & \textbf{ROUGE-2} & \textbf{ROUGE-L} \\
\midrule
ChatGPT-L & 38.96 & 20.53 & 35.86 \\
ChatGPT-S & 39.28 & 21.89 & 32.74 \\
Vicuna & 62.51 & 53.98 & 61.95 \\
Vicuna-FT & \textbf{33.39} & \textbf{20.00} & \textbf{31.45}\\
\bottomrule
\end{tabular}%
}
\caption{Diversity of doctors' query. The table shows three types of diversity scores with different metrics. Note that a lower score represents more distinction between the models' questions.}
\label{tab: diversity}
\end{table}


It can be observed that the accuracy and F1 scores have an overall increasing trend as the dialogue progresses. The indicators quickly improve during the initial rounds of the conversation and then tend to plateau. This suggests that relevant information is easily acquired at the beginning of the consultation, while the challenge lies in obtaining further details of the patient's information. Among the models, Vicuna-FT has the highest F1 score for both doctor and patient, indicating that it has less hallucination problem and can obtain more information from patients, resulting in the best overall performance in accuracy score.


Note that there is a slight drop in both doctor F1 score and accuracy after 80$\%$ of the dialogue process. This is because the doctor LLMs are prone to give a diagnosis or advice near the end of the consultation process, which adds noise to the dialogue history and causes performance degradation in the final task.


\begin{figure}[thbp]
\centering
\includegraphics[width=1\columnwidth]{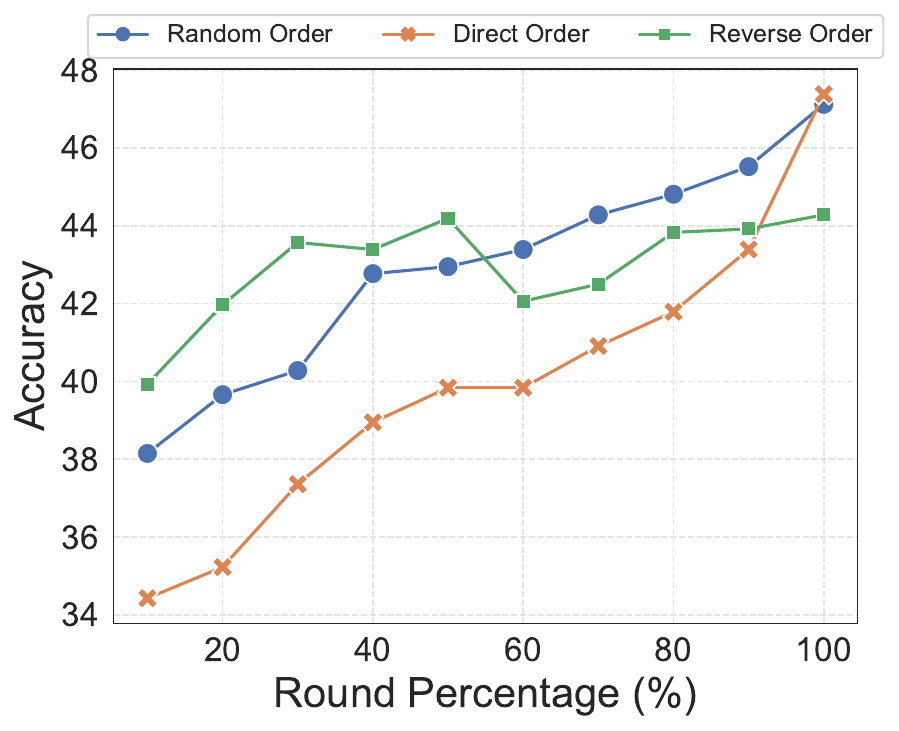}
\caption{Accuracy curves of random, direct, and reversed orders of medical information keys.}
\label{fig:oracle_analysis}
\end{figure}

\subsection{Diversity of Doctors' Queries}
There are some repetition queries found in the case study. Therefore, we calculated the similarity between the queries of doctor LLMs in the same conversation as follows:
\begin{equation}
    {\rm Diversity} = \frac{1}{K(K-1)}\sum^K_{i=1}\sum^K_{j=1}{\rm sim}(D_i, D_j)
\end{equation}
where $K$ is the total rounds of the consultation conversation, and ${\rm sim}(\cdot, \cdot)$ is the similarity function. As presented in Table~\ref{tab: diversity}, Vicuna-FT generates the most diverse queries among all models, while Vicunas generate the most similar ones. Additionally, we can also observe that Vicuna tended to repeat questions during the consultation process.

\subsection{Impact of Consultation Order}
We also conducted experiments to analyze the impact of the order of doctor consultation. To eliminate the effect of different questions, we conducted an oracle case where the doctor's questions are constructed by the keys of the extracted medical information. Specifically, at each turn, we formulated the rule-based question in the form of "Can you tell me about $\{\{$key$\}\}$?" according to the (key, value) pair in the extracted medical information list.

Since the keys in the extracted medical information typically range from basic medical information to more specialized medical details, we can construct three types of consultation orders, which includes direct order, random order, and reverse order. Direct order is to ask medical information questions from general to specific, while the reverse order is exactly the opposite. Figure~\ref{fig:oracle_analysis} shows their impacts on final accuracy.

In the first part of the conversation, the reversed order performed the best, indicating that detailed information is more helpful in solving the final task. However, the final accuracy of the reverse case is lower than the other two settings even though all the cases have equal patient information. We attribute this phenomenon to the task solver's sensitivity to the location of patient information in the context.


\begin{figure}[thbp]
\centering
\includegraphics[width=1.0\linewidth]{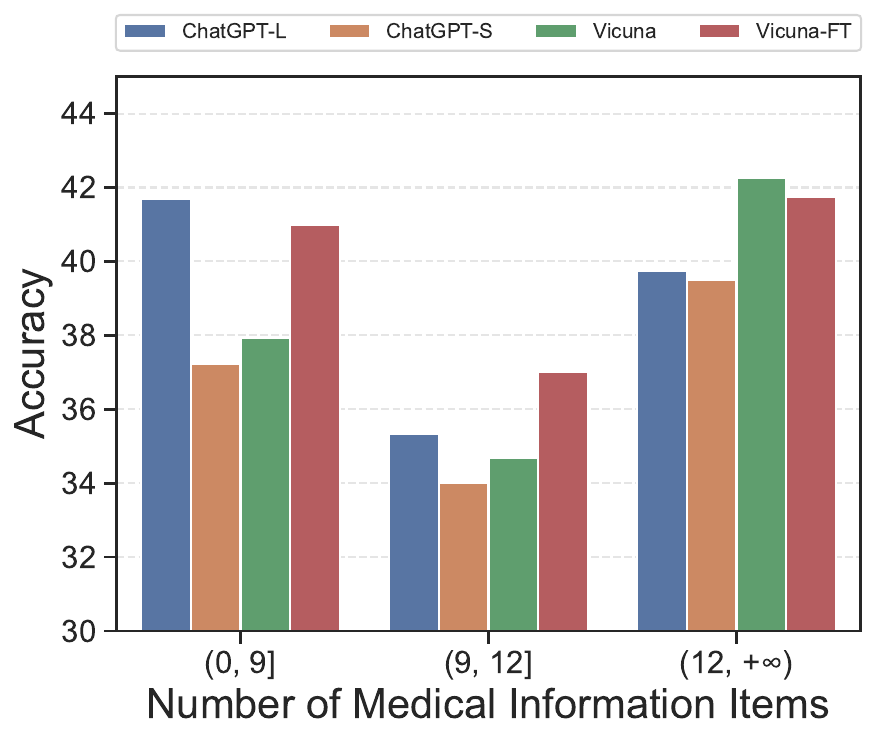}
\caption{The performance of different models with respect to the number of medical information items. The results are reported on MedQA.}
\label{fig:barplot}
\end{figure}

\subsection{Performance on Data with Different Complexity}
\label{sec:acc}

In this section, we investigate the effectiveness of different models on data with varying numbers of medical information items. The entire test set is divided into three equal parts based on the number of medical information items. As shown in Figure~\ref{fig:barplot}, all models perform poorly on data with a moderate amount of medical information. However, Vicuna-FT exhibits robust performance across all three groups. It is worth noting that ChatGPT-Long outperforms Vicuna on the Short group, while Vicuna outperforms ChatGPT-Long on the Long group. This indicates that models have a bias towards certain numbers of medical information items.

\section{Conclusions}

In this paper, we proposed a novel framework to evaluate the medical ability of LLMs in multi-turn consultations and construct three test datasets by reformulating the multiple-choice questions. We adopt ChatGPT to simulate the patient in the consultation process in order to generate responses to the question of doctor LLMs. Besides, we develop comprehensive evaluation metrics to score the multi-turn conversation and improve the consultation ability and alleviate the hallucination of LLMs by constructing a training set. We also conduct extensive experiments to valid the robustness of our evaluation pipeline.

\section*{Limitations}

Although we have utilized the most superior datasets available, this study may still be limited due to the size and potential bias of said datasets. Furthermore, the limited quantity of medical examination examples collected may hinder our ability to evaluate the model's comprehensive capabilities in a real-world consultation scenario.
Moreover, it is important to note that LLMs still encounter challenges related to hallucination and robustness, which may make the task solver susceptible to long and noisy inputs. Thus, in order to construct a more dependable and resilient solution, a more powerful LLM is necessary.

\bibliography{anthology,custom}

\begin{thebibliography}{19}
\expandafter\ifx\csname natexlab\endcsname\relax\def\natexlab#1{#1}\fi

\bibitem[{Agrawal et~al.(2022)Agrawal, Hegselmann, Lang, Kim, and
  Sontag}]{DBLP:conf/emnlp/AgrawalHLKS22}
Monica Agrawal, Stefan Hegselmann, Hunter Lang, Yoon Kim, and David~A. Sontag.
  2022.
\newblock \href {https://aclanthology.org/2022.emnlp-main.130} {Large language
  models are few-shot clinical information extractors}.
\newblock In \emph{Proceedings of the 2022 Conference on Empirical Methods in
  Natural Language Processing, {EMNLP} 2022, Abu Dhabi, United Arab Emirates,
  December 7-11, 2022}, pages 1998--2022. Association for Computational
  Linguistics.

\bibitem[{Chen et~al.(2023)Chen, Wu, Zhu, Lan, Zhang, and
  Cui}]{DBLP:journals/corr/abs-2305-13614}
Siyuan Chen, Mengyue Wu, Kenny~Q. Zhu, Kunyao Lan, Zhiling Zhang, and Lyuchun
  Cui. 2023.
\newblock \href {https://doi.org/10.48550/arXiv.2305.13614} {Llm-empowered
  chatbots for psychiatrist and patient simulation: Application and
  evaluation}.
\newblock \emph{CoRR}, abs/2305.13614.

\bibitem[{Han et~al.(2023)Han, Adams, Papaioannou, Grundmann, Oberhauser,
  L{\"{o}}ser, Truhn, and Bressem}]{DBLP:journals/corr/abs-2304-08247}
Tianyu Han, Lisa~C. Adams, Jens{-}Michalis Papaioannou, Paul Grundmann, Tom
  Oberhauser, Alexander L{\"{o}}ser, Daniel Truhn, and Keno~K. Bressem. 2023.
\newblock \href {https://doi.org/10.48550/arXiv.2304.08247} {Medalpaca - an
  open-source collection of medical conversational {AI} models and training
  data}.
\newblock \emph{CoRR}, abs/2304.08247.

\bibitem[{Hiesinger et~al.(2023)Hiesinger, Zakka, Chaurasia, Shad, Dalal, Kim,
  Moor, Alexander, Ashley, Boyd et~al.}]{hiesinger2023almanac}
William Hiesinger, Cyril Zakka, Akash Chaurasia, Rohan Shad, Alex Dalal,
  Jennifer Kim, Michael Moor, Kevin Alexander, Euan Ashley, Jack Boyd, et~al.
  2023.
\newblock Almanac: Retrieval-augmented language models for clinical medicine.

\bibitem[{Kung et~al.(2023)Kung, Cheatham, Medenilla, Sillos, De~Leon,
  Elepaño, Madriaga, Aggabao, Diaz-Candido, Maningo, and
  Tseng}]{10.1371/journal.pdig.0000198}
Tiffany~H. Kung, Morgan Cheatham, Arielle Medenilla, Czarina Sillos, Lorie
  De~Leon, Camille Elepaño, Maria Madriaga, Rimel Aggabao, Giezel
  Diaz-Candido, James Maningo, and Victor Tseng. 2023.
\newblock \href {https://doi.org/10.1371/journal.pdig.0000198} {Performance of
  chatgpt on usmle: Potential for ai-assisted medical education using large
  language models}.
\newblock \emph{PLOS Digital Health}, 2(2):1--12.

\bibitem[{Li et~al.(2023)Li, Li, Zhang, Dan, and
  Zhang}]{DBLP:journals/corr/abs-2303-14070}
Yunxiang Li, Zihan Li, Kai Zhang, Ruilong Dan, and You Zhang. 2023.
\newblock \href {https://doi.org/10.48550/arXiv.2303.14070} {Chatdoctor: {A}
  medical chat model fine-tuned on llama model using medical domain knowledge}.
\newblock \emph{CoRR}, abs/2303.14070.

\bibitem[{Lin(2004)}]{lin-2004-rouge}
Chin-Yew Lin. 2004.
\newblock \href {https://aclanthology.org/W04-1013} {{ROUGE}: A package for
  automatic evaluation of summaries}.
\newblock In \emph{Text Summarization Branches Out}, pages 74--81, Barcelona,
  Spain. Association for Computational Linguistics.

\bibitem[{Liu et~al.(2022)Liu, Tang, Cheng, Li, Zheng, and
  Liang}]{DBLP:conf/nlpcc/LiuTCLZL22}
Wenge Liu, Jianheng Tang, Yi~Cheng, Wenjie Li, Yefeng Zheng, and Xiaodan Liang.
  2022.
\newblock \href {https://doi.org/10.1007/978-3-031-17120-8\_35} {Meddg: An
  entity-centric medical consultation dataset for entity-aware medical dialogue
  generation}.
\newblock In \emph{Natural Language Processing and Chinese Computing - 11th
  {CCF} International Conference, {NLPCC} 2022, Guilin, China, September 24-25,
  2022, Proceedings, Part {I}}, volume 13551 of \emph{Lecture Notes in Computer
  Science}, pages 447--459. Springer.

\bibitem[{Luo et~al.(2022)Luo, Sun, Xia, Qin, Zhang, Poon, and
  Liu}]{DBLP:journals/bib/LuoSXQZPL22}
Renqian Luo, Liai Sun, Yingce Xia, Tao Qin, Sheng Zhang, Hoifung Poon, and
  Tie{-}Yan Liu. 2022.
\newblock \href {https://doi.org/10.1093/bib/bbac409} {Biogpt: generative
  pre-trained transformer for biomedical text generation and mining}.
\newblock \emph{Briefings Bioinform.}, 23(6).

\bibitem[{Nori et~al.(2023)Nori, King, McKinney, Carignan, and
  Horvitz}]{DBLP:journals/corr/abs-2303-13375}
Harsha Nori, Nicholas King, Scott~Mayer McKinney, Dean Carignan, and Eric
  Horvitz. 2023.
\newblock \href {https://doi.org/10.48550/arXiv.2303.13375} {Capabilities of
  {GPT-4} on medical challenge problems}.
\newblock \emph{CoRR}, abs/2303.13375.

\bibitem[{OpenAI(2022)}]{elmohamed}
OpenAI. 2022.
\newblock Chatgpt: Optimizing language models for dialogue.
\newblock Website.
\newblock \url{https://openai.com/blog/chatgpt}.

\bibitem[{OpenAI(2023)}]{DBLP:journals/corr/abs-2303-08774}
OpenAI. 2023.
\newblock \href {https://doi.org/10.48550/arXiv.2303.08774} {{GPT-4} technical
  report}.
\newblock \emph{CoRR}, abs/2303.08774.

\bibitem[{Singhal et~al.(2023)Singhal, Tu, Gottweis, Sayres, Wulczyn, Hou,
  Clark, Pfohl, Cole{-}Lewis, Neal, Schaekermann, Wang, Amin, Lachgar,
  Mansfield, Prakash, Green, Dominowska, y~Arcas, Tomasev, Liu, Wong, Semturs,
  Mahdavi, Barral, Webster, Corrado, Matias, Azizi, Karthikesalingam, and
  Natarajan}]{DBLP:journals/corr/abs-2305-09617}
Karan Singhal, Tao Tu, Juraj Gottweis, Rory Sayres, Ellery Wulczyn, Le~Hou,
  Kevin Clark, Stephen Pfohl, Heather Cole{-}Lewis, Darlene Neal, Mike
  Schaekermann, Amy Wang, Mohamed Amin, Sami Lachgar, Philip~Andrew Mansfield,
  Sushant Prakash, Bradley Green, Ewa Dominowska, Blaise~Ag{\"{u}}era y~Arcas,
  Nenad Tomasev, Yun Liu, Renee Wong, Christopher Semturs, S.~Sara Mahdavi,
  Joelle~K. Barral, Dale~R. Webster, Gregory~S. Corrado, Yossi Matias,
  Shekoofeh Azizi, Alan Karthikesalingam, and Vivek Natarajan. 2023.
\newblock \href {https://doi.org/10.48550/arXiv.2305.09617} {Towards
  expert-level medical question answering with large language models}.
\newblock \emph{CoRR}, abs/2305.09617.

\bibitem[{Touvron et~al.(2023)Touvron, Lavril, Izacard, Martinet, Lachaux,
  Lacroix, Rozi{\`{e}}re, Goyal, Hambro, Azhar, Rodriguez, Joulin, Grave, and
  Lample}]{DBLP:journals/corr/abs-2302-13971}
Hugo Touvron, Thibaut Lavril, Gautier Izacard, Xavier Martinet, Marie{-}Anne
  Lachaux, Timoth{\'{e}}e Lacroix, Baptiste Rozi{\`{e}}re, Naman Goyal, Eric
  Hambro, Faisal Azhar, Aur{\'{e}}lien Rodriguez, Armand Joulin, Edouard Grave,
  and Guillaume Lample. 2023.
\newblock \href {https://doi.org/10.48550/arXiv.2302.13971} {Llama: Open and
  efficient foundation language models}.
\newblock \emph{CoRR}, abs/2302.13971.

\bibitem[{Wei et~al.(2018)Wei, Liu, Peng, Tou, Chen, Huang, Wong, and
  Dai}]{DBLP:conf/acl/WeiLPTCHWD18}
Zhongyu Wei, Qianlong Liu, Baolin Peng, Huaixiao Tou, Ting Chen, Xuanjing
  Huang, Kam{-}Fai Wong, and Xiangying Dai. 2018.
\newblock \href {https://doi.org/10.18653/v1/P18-2033} {Task-oriented dialogue
  system for automatic diagnosis}.
\newblock In \emph{Proceedings of the 56th Annual Meeting of the Association
  for Computational Linguistics, {ACL} 2018, Melbourne, Australia, July 15-20,
  2018, Volume 2: Short Papers}, pages 201--207. Association for Computational
  Linguistics.

\bibitem[{Zeng et~al.(2020)Zeng, Yang, Ju, Yang, Wang, Zhang, Zhou, Zeng, Dong,
  Zhang, Fang, Zhu, Chen, and Xie}]{DBLP:conf/emnlp/ZengYJYWZZZDZFZ20}
Guangtao Zeng, Wenmian Yang, Zeqian Ju, Yue Yang, Sicheng Wang, Ruisi Zhang,
  Meng Zhou, Jiaqi Zeng, Xiangyu Dong, Ruoyu Zhang, Hongchao Fang, Penghui Zhu,
  Shu Chen, and Pengtao Xie. 2020.
\newblock \href {https://doi.org/10.18653/v1/2020.emnlp-main.743} {Meddialog:
  Large-scale medical dialogue datasets}.
\newblock In \emph{Proceedings of the 2020 Conference on Empirical Methods in
  Natural Language Processing, {EMNLP} 2020, Online, November 16-20, 2020},
  pages 9241--9250. Association for Computational Linguistics.

\bibitem[{Zhang et~al.(2023{\natexlab{a}})Zhang, Chen, Jiang, Yu, Chen, Li,
  Chen, Wu, Zhang, Xiao, Wan, Wang, and Li}]{DBLP:journals/corr/abs-2305-15075}
Hongbo Zhang, Junying Chen, Feng Jiang, Fei Yu, Zhihong Chen, Jianquan Li,
  Guiming Chen, Xiangbo Wu, Zhiyi Zhang, Qingying Xiao, Xiang Wan, Benyou Wang,
  and Haizhou Li. 2023{\natexlab{a}}.
\newblock \href {https://doi.org/10.48550/arXiv.2305.15075} {Huatuogpt, towards
  taming language model to be a doctor}.
\newblock \emph{CoRR}, abs/2305.15075.

\bibitem[{Zhang et~al.(2023{\natexlab{b}})Zhang, Press, Merrill, Liu, and
  Smith}]{DBLP:journals/corr/abs-2305-13534}
Muru Zhang, Ofir Press, William Merrill, Alisa Liu, and Noah~A. Smith.
  2023{\natexlab{b}}.
\newblock \href {https://doi.org/10.48550/arXiv.2305.13534} {How language model
  hallucinations can snowball}.
\newblock \emph{CoRR}, abs/2305.13534.

\bibitem[{Zheng et~al.(2023)Zheng, Huang, and
  Chang}]{DBLP:journals/corr/abs-2304-10513}
Shen Zheng, Jie Huang, and Kevin~Chen{-}Chuan Chang. 2023.
\newblock \href {https://doi.org/10.48550/arXiv.2304.10513} {Why does chatgpt
  fall short in answering questions faithfully?}
\newblock \emph{CoRR}, abs/2304.10513.

\end{thebibliography}
\bibliographystyle{acl_natbib}

\appendix

\end{document}